\pdfoutput=1
\def\year{2019}\relax
\documentclass[letterpaper]{article} %DO NOT CHANGE THIS
\usepackage{aaai19}  %Required
\usepackage{times}  %Required
\usepackage{helvet}  %Required
\usepackage{courier}  %Required
\usepackage{url}  %Required
\usepackage{graphicx}  %Required
\usepackage{subcaption}
\usepackage{amsmath}
\usepackage{color}
\begin{document}
% The submission should be anonymous
\author{Ziyuan Pan,\textsuperscript{1}
Hao Du,\textsuperscript{2, *}
Kee Yuan Ngiam,\textsuperscript{3}
Fei Wang, \textsuperscript{4}
Ping Shum,\textsuperscript{1}
Mengling Feng,\textsuperscript{2}
\\
\textsuperscript{1}{School of Electrical and Electronic Engineering, Nanyang Technological University, Singapore 639798}\\
\textsuperscript{2}{Saw Swee Hock School of Public Health, National University Health System, National University of Singapore, Singapore 117549}\\
\textsuperscript{3}{National University Health System, Singapore 119228}\\
\textsuperscript{4}{Healthcare Policy and Research, Weill Cornell Medical College, Cornell University, New York, NY 10065}\\
\textsuperscript{*}{Corresponding author email: duhao@u.nus.edu}
}
\title{A Self-Correcting Deep Learning Approach to Predict Acute Conditions in Critical Care}
\maketitle
\begin{abstract}
In critical care, intensivists are required to continuously monitor high dimensional vital signs and lab measurements to detect and diagnose acute patient conditions. This has always been a challenging task. In this study, we propose a novel self-correcting deep learning prediction approach to address this challenge. We focus on an example of the prediction of acute kidney injury (AKI). Compared with the existing models, our method has a number of distinct features: we utilized the accumulative data of patients in ICU; we developed a self-correcting mechanism that feeds errors from the previous predictions back into the network; we also proposed a regularization method that takes into account not only the model's prediction error on the label but also its estimation errors on the input data. This mechanism is applied in both regression and classification tasks. We compared the performance of our proposed method with the conventional deep learning models on two real-world clinical datasets and demonstrated that our proposed model constantly outperforms these baseline models. In particular, the proposed model achieved area under ROC curve at 0.893 on the MIMIC III dataset, and 0.871 on the Philips eICU dataset.%, and we obtained AUROC at 0.871. Inspired by the promising outcome, we plan to further validate the proposed method with local data from our hospital with the ultimate goal of deploying it as a clinical decision support tool.
\end{abstract}

%%%%%%%%%%%%%%%%%%%%%%%%%%%%%%%%%%%
\section{Introduction}
\noindent Electronic Health Records (EHR) data are accumulative, routinely collected patient observations from hospitals or clinical institutes. In the case of intensive care unit (ICU), EHR include not only the static information such as patient demographics, discrete time-series data such as medication and diagnosis, but also continuous multi-variate time-series data such as vital signs and laboratory measurements. 
In order to detect and diagnose acute (and usually deadly) patient conditions, ICU intensivists need to continuously monitor high dimensional vital signs and lab measurements \cite{Harty2014Prevention}. Example acute conditions include acute kidney injury, acute hypertension, acute organ failure and acute septic shocks. It has always been challenging to track all indicative changes in various patients' data to quickly diagnose these acute conditions. Predictive models developed with ICU EHR data provides an opportunity for early detection of ICU acute conditions, which can lead to in-time and better care. In this study, we propose a self-correcting deep learning framework with accumulative ICU data to predict these acute conditions. Without the loss of generality, we will focus on the prediction of acute kidney injury (AKI), while the predictive modeling of other conditions can be proceeded similarly.

AKI is a sudden onset of renal failure or kidney damage, occurring in at least 5\% of hospitalized patients. It is associated with a significant increase on mortality, length of stay (LOS) and hospital cost under a wide range of conditions \cite{chertow2005acute}. AKI is a very good study case for disease risk predictive modeling because: 1) the precise definition for AKI allows temporal anchoring of events; 2) if detected and managed in time, AKI is potentially avoidable and reversible in the process of a few hours to several days \cite{kate2016prediction}. An accurate, automated and early AKI detection system would prevent AKI events, thus reducing mortality, shortening LOS, avoiding the development of chronic kidney disease and potentially creating quality of care indicators \cite{sutherland2016utilizing}.

Existing study from Kate et al. demonstrated early prediction of AKI using multiple machine learning methods, including logistic regression, support vector machine, decision trees and naive Bayes, on a population of hospitalized older adults \cite{kate2016prediction}. The models were based on patients' demographic information, comorbidities, family history, medications and laboratory values in EHRs. For each patient, the models only used the last recorded value before 24 h after admission. The time dependency in EHR data is not captured by the models.
In the setting of ICU, or more in general the setting of in-patient care, after patients' admission, their situation often evolve over time rapidly. Therefore, patients' EHR data is a dynamic time-series in nature. 
The challenges of time-dependent EHR data, including event temporality, high-dimensionality and irregular sampling, have been investigated in recent years using a series of machine learning methods, especially recurrent neural networks (RNNs) \cite{qiao2018pairwise}.  
Lipton et al. demonstrated utilizing RNNs in ICU diagnoses \cite{lipton2015learning} and showed that RNNs are capable of capturing time dependencies between the elements. In recent studies, researchers demonstrated utilization of RNNs or convolutional neural networks (CNNs) to improve the prediction of heart failure onsets \cite{choi2016doctor,cheng2016risk}. In \cite{ma2018health}, the challenges were addressed by employing RNNs with attention mechanism and add interpretability of the prediction results. 
In addition to these challenges, patients' EHR data also have an accumulative characteristics: as patients stay longer in the hospital, more data are collected about their disease progression, and thus a more accurate modeling of patients' physiological states is possible.  
Conventional RNNs were not optimized to accumulate information in time series data. The accumulated error between prediction and the patient's status are not specifically modeled across the patient's ICU stay. 
In addition, an effective ICU acute condition predictor is expected to enhance itself through self-correcting and learning from its accumulated prediction errors. This self-correcting mechanism is lacking in conventional RNN models.

To address the above limitations, we propose a variant of RNN to predict the onset of patients' AKI in ICU. 
In this pilot study, we validate the effectiveness of our proposed self-correcting model with two actual ICU patient EHR datasets from the US. In the next phase, we plan to validate and deploy our algorithm locally in our own hospital.

The main contributions of this study are summarized as follows.
\begin{enumerate}
  \item Our method utilized the accumulative data of patients in ICU instead of a snapshot of the patient's condition to improve the performance of AKI prediction. 
  \item We developed a novel accumulative self-correcting mechanism by modeling the accumulated errors in the model when the prediction is incorrect. 
  \item We proposed a regularization method for our model, which takes into account not only the model's prediction error on the label but also its estimation errors on the future input data. Such regularization reduces the variance of the model and improves the efficiency of the self-correcting mechanism.
  \item Our proposed method has been validated in two real-world large scale ICU datasets. It was shown to outperform traditional RNNs. In addition, the method is in-progress of being validated locally with data from our own hospital.
%NUH has developed a platform, named DISCOVERY AI \cite{stimes}, to allow machine learning algorithms to be easily validated and deployed in the actual clinical settings. With the demonstrated results, further deployment and improvement are in progress for future work. 
%  \item Our method is capable to model an AKI patient's trajectory in ICU, including his/her AKI onset and AKI staging, which would be useful for physicians to provide fast response and care. 

\end{enumerate}

%%%%%%%%%%%%%%%%%%%%%%%%%%%%%%%%%%%%%
\section{Related Works}
AKI prediction with utilization of features from EHR data is attracting a widespread research interest \cite{sutherland2016utilizing,weisenthal2017sum}. In particular, much research in recent years has focused on predictive modeling on a broad population to identify high-risk subjects as early as possible \cite{weisenthal2017sum}. Initially, AKI prediction was modeled by standard statistical modeling methods, including logistic regression, discriminant analysis, or decision tree algorithm \cite{thakar2005clinical,palomba2007acute,brown2007multivariable,mohamadlou2018prediction}. Data were accumulated using sliding window method, and the prediction was generated at a specified interval (per hour, two hours, day, shift etc). Alternatively, some models \cite{sutherland2016utilizing} could
generate a risk score in real time when a new data point was received as well.  

Recently, a number of studies have been carried out utilizing recurrent neural networks (RNNs) in clinical diagnosis and prediction. RNNs are one branch of neural networks, which are powerful to process sequential data \cite{miotto2017deep}. 
%
% Recently, RNNs have been successfully employed in natural language processing \cite{wen2015semantically}, speech recognition \cite{graves2013speech}, text classification \cite{lai2015recurrent} and video processing \cite{donahue2015long}. 
%
In RNNs, hidden units connect to each other by forming a directed cycle. Each output value is dependent on the previous computations. In traditional RNNs, the network can only look back to a few steps due to the gradient vanishing and exploding problems \cite{miotto2017deep}. To address these limitations, variants of RNNs, such as LSTM \cite{hochreiter1997long} and GRU \cite{cho2014learning}, are proposed and well utilized in clinical prediction problems. These variants model have the hidden state with the forget gate that decide what to keep in and erase from the memory of the network. 

Lipton et al. presented the first study to empirically evaluate LSTMs on pattern recognition in multivariate clinical time series data \cite{lipton2015learning}. The authors employed multilabel LSTM to classify 128 diagnoses given 13 frequently but irregularly sampled clinical measurements in in pediatric intensive unit care. Compared with several strong baselines, including a multilayer perceptron trained on hand-engineered features, LSTM showed significant improvement in accuracy. 
Gated recurrent unit (GRU) were then proposed by Choi et al. \cite{choi2016doctor} to develop Doctor AI, a temporal predictive model that accesses the longitudinal time stamped EHR data of patients to predict the diagnosis and medication categories for a subsequent visit. Their method significantly outperformed shallow baselines with higher recall and proved that not only diagnosis, the disease progression can be well captured by RNNs. 
In addition to LSTMs, pooling and word embedding were used in DeepCare \cite{pham2016deepcare} to model illness states of patients and to predict patients' outcomes. DeepCare is an end-to-end deep dynamic memory neural network. DeepCare introduced time parameterizations to handle irregular timed events and utilized accumulative temporal data by moderating the forgetting and consolidation of memory cells. 
%
%Moreover, the proposed method incorporated medical interventions that change the course of illness to dynamically shape future risks. 
%
DeepCare demonstrated improved disease progression modeling accuracy and risk prediction compared to Markov models and plain RNNs. 
The limitation that is shared by both Doctor AI and DeepCare is that, as they continue to predict patients' disease progression, they lack a feedback mechanism to allow the model to learn and improve from its previous prediction mistakes.

% Bidirectional RNNs were also employed for diagnosis prediction in \cite{ma2017dipole}. 
% %
% Bidirectional RNNs can be trained by using all the available input temporal data from two directions to improve the prediction performance \cite{graves2013hybrid,chiu2015named}. 
% However, their prediction performance may drop significantly as the length of the sequential data gets longer \cite{schuster1997bidirectional}. 

%In , Ma et al. demonstrated attention-based bidirectional RNNs outperformed baseline methods in diagnosis prediction. The proposed method were designed to remember all the information from both past and future visits and it introduced attention mechanisms to inteprete the relationships between differenct visits for the prediction. 

Specifically on AKI prediction, a number of studies have been carried out. Hurry et al. employed Bayesian Networks on AKI prediction by predicting
the likelihood of AKI onset based on longitudinal
patient data on MIMIC II database \cite{cruz12early}. In addition, Nogueira et al. applied Markov Chain model on PhysioNet dataset to predict the future state of the patients based on the current medical state and ICU type. The common limitation of the these studies is that the proposed methods are all dependent on hand-engineered features and expert knowledge, where the hidden states of the patient condition were not effectively modeled. 

To address the mentioned limitations in previous works, we proposed a novel RNNs based method to predict acute condition in ICU. Our method introduced a self-correcting mechanism coupled with a regularization method to further optimize the prediction error. 

\section{METHODOLOGY}
\subsection{Problem Definition}
\begin{figure}
\centering
\includegraphics[width=.8\linewidth]{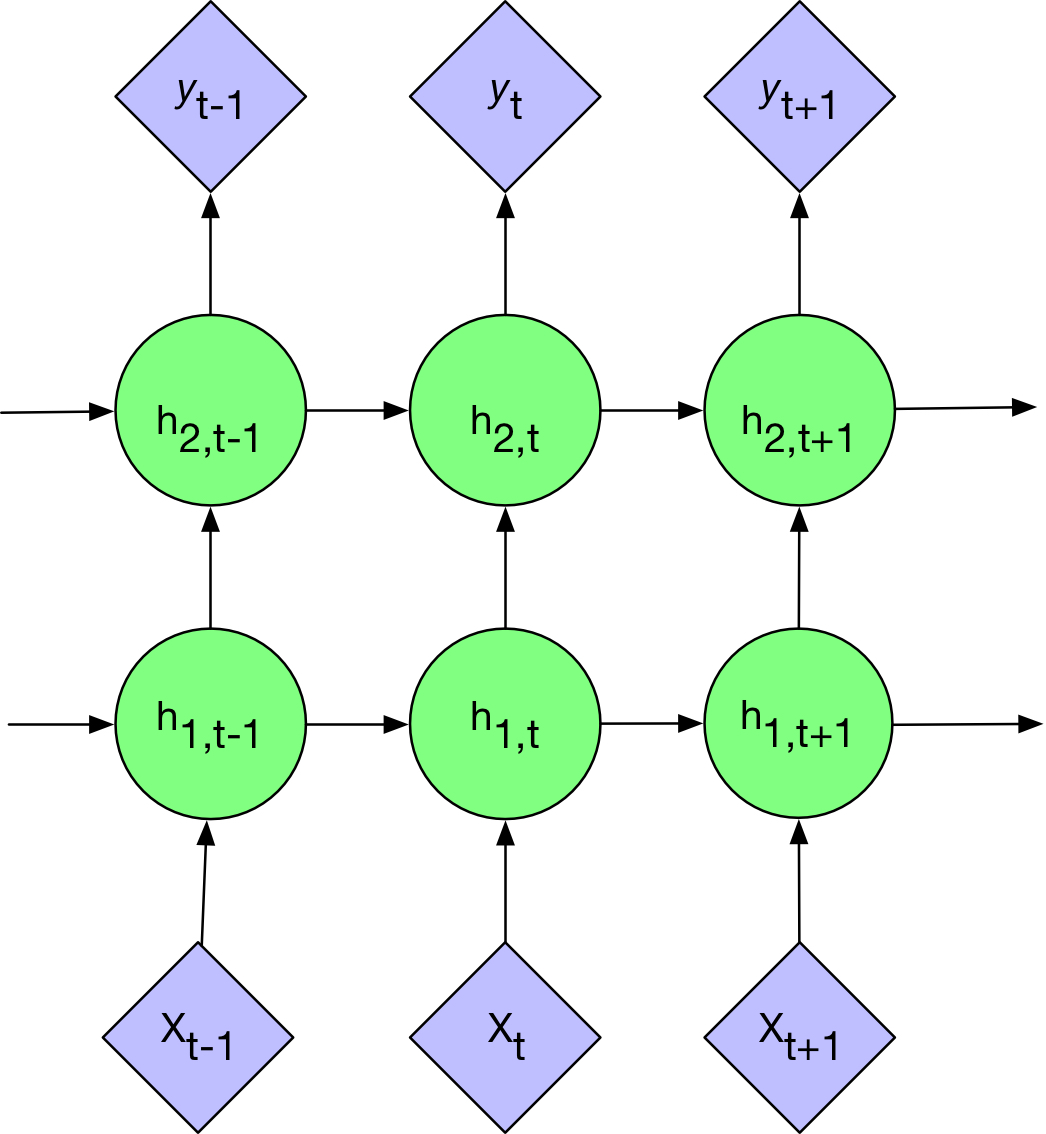}
\caption{A diagram of a multi-layer RNN.}
\label{fig:RNN}
\end{figure}

For any ICU patient and any time point $t$ during his/her ICU stay, our goal is to predict whether the ICU patient may develop AKI in next 6 hours, i.e. $t+6$, based on all his/her data of this ICU stay accumulated up till time $t$ .
%, based on all existing temporal information. 
%
%For classifying patients on AKI onset, considering the severity of their state and act according to that, there are several criteria to be applied. 
%
In this study, AKI was defined according to the most commonly used RIFLE criteria \cite{bellomo2004acute}. 
A patient was detected with AKI if his/her urine output is less than 0.5mL/kg/h for $\geq$ 6h\\
Based on this definition, for a patient, the AKI actual onset label at any time $t$, denoted as $y_t$, can only be observed at time step $t+6$. In the traditional RNN (shown in Figure~\ref{fig:RNN}), the correctness of the predicted $\hat{y}_t$ would not affect the prediction in the future time step, although at time step t+6, we will know whether our predicted  $\hat{y}_t = y_t$. In this study, we want to fully utilize all the observed data including the label $y_t$ in order to continuously improve the accuracy of the model. Therefore, we designed the novel self-correcting mechanism to further enhance the conventional RNN model.

\subsection{Data and Data Preprocessing}
We applied our proposed method  on the Medical Information Mart for Intensive Care III (MIMIC-III) and Phillips eICU Collaborative Research Dataset. MIMIC-III dataset \cite{johnson2016mimic} consists of medical records of over 40,000 ICU patients between 2001 and 2012. Data in MIMIC-III include demographic information, vital signs, medication records, laboratory measurements, observations, fluid balance, procedure codes, diagnostic codes, imaging reports, hospital length of stay, survival data, and so on. We further validated the performance of our proposed method with the Phillips eICU Collaborative Research Dataset. The eICU dataset is populated with data from a combination of multiple ICU across the United States \cite{moody2001physionet}. The dataset covers 20,0859 ICU patients admitted in 2014 and 2015.

For this project, we extracted the following variables from both the MIMIC-III and eICU datasets:

\begin{enumerate}
\item \textbf{Demographic information} (\textit{static variables}): Age and Gender
\item \textbf{Co-morbidities} (\textit{static variables}): ICD-9 \cite{american2004international} defined co-morbidities conditions of Congestive Heart Failure, Cardiac Arrhythmias, Valvular Disease, Pulmonary Circulation, Peripheral Vascular, Hypertension, Paralysis, Neurological Disorder, Chronic Pulmonary Diseases, Diabetes, Hypothyroidism, Renal Failure, Liver Diseases, Peptic Ulcer, AIDS, Lymphoma, Metastatic Cancer, Rheumatoid Arthritis, Coagulopathy, Obesity, Fluid Electrolyte, Anemias, Alcohol Abuse, Drug Abuse, Psychoses and Depression. 
\item \textbf{Vital signs} (\textit{time-series variables}): Mean Arterial Blood Pressure, Heart Rate, Respiration Rate, Temperature
\item \textbf{Lab measurements} (\textit{time-series variables}): 
Bilirubin,
BUN (Blood urea nitrogen),
Creatinine,
Glucose,
HCO3 (Serum bicarbonate),
HCT (Hematocrit),
K (Serum potassium),
Lactate,
Mg (Serum magnesium),
Na (Serum sodium),
PaCO2 (partial pressure of arterial CO2,
PaO2 (Partial pressure of arterial O2,
pH,
Platelets,
Troponin,
WBC (White Blood Cell count).
\item \textbf{Fluids} (\textit{time-series variables}): Urine Output and Fluid Balance
\item \textbf{Interventions} (\textit{time-series variables}): Usage of Mechanical Ventilation, Vasopressor and sedative medications.

\end{enumerate}

For the extracted time-series variables, the vital signs were regularly collected on hourly basis. But the lab measurements, fluids information and interventions were collected with random time windows. For these variables, we transformed the data into regularly sample time series, where the time gap between two data point is always one hour. For the time steps where there was no recorded data, data were imputed with the weighted average value of the nearest data points.

\subsection{Proposed Method}
\subsubsection{General Idea}

\begin{figure}[h!]
\centering
\includegraphics[width=1.0\linewidth]{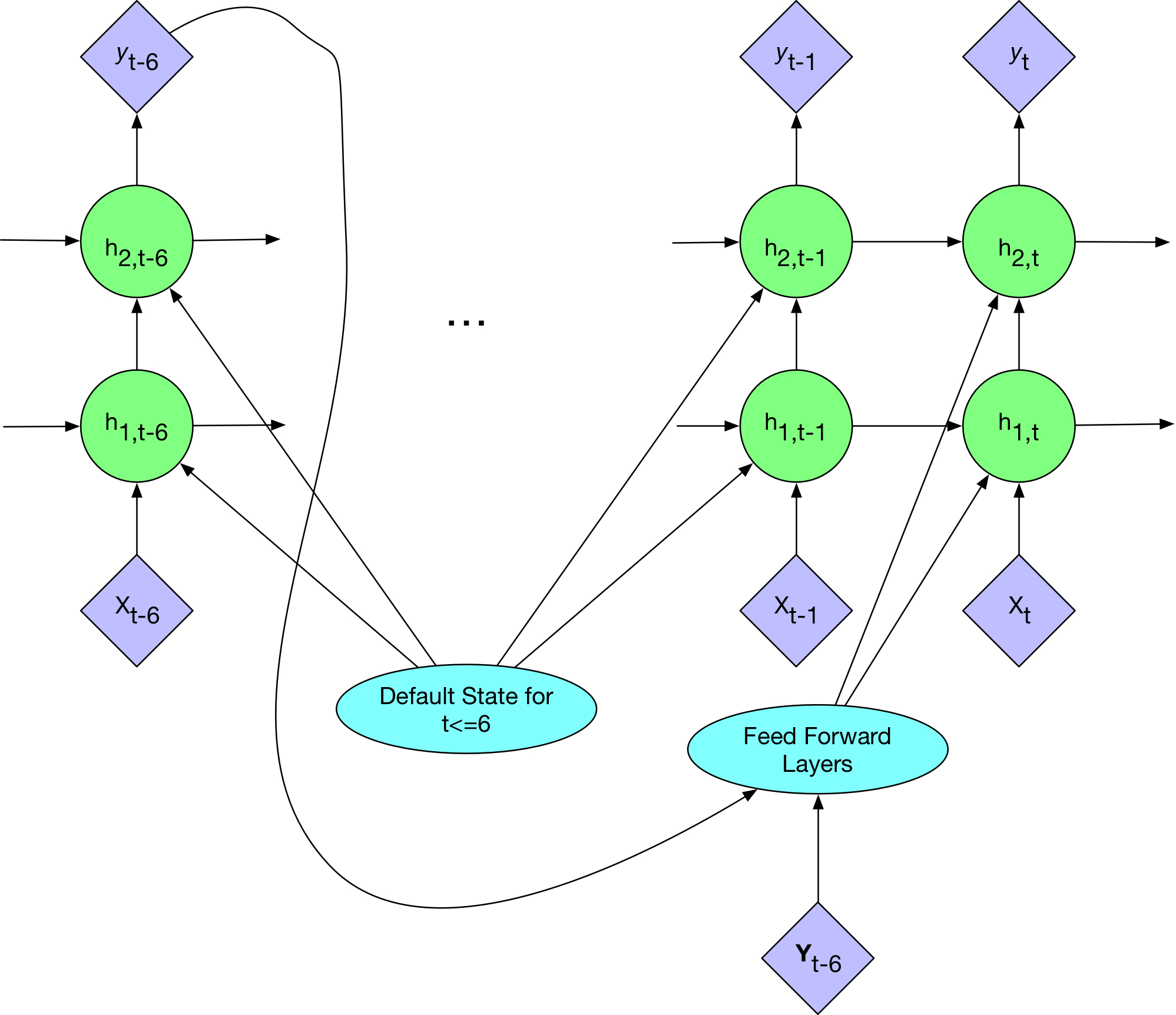}
\caption{\label{fig:Self-correcting_RNN}RNN with self-correcting mechanism. Circles are used for RNN cells (either LSTM or GRU), while diamond shaped units are used for input and output. Italic letters (e.g. \textit{$x_t$, $y_t$}) denote the predicted values, while bold capital letters (e.g. \textbf{$X_t$, $Y_t$}) denote the actual values.}
\end{figure}

Figure~\ref{fig:Self-correcting_RNN} graphically illustrated the proposed Self-correcting RNN framework. Compared to the traditional multi-layer RNN, we created a feedback loop between each time step t and t-6, $\forall t \in \{7...T\}$. At each time step $t$, we have $\hat{y}_{t-6}$,  representing the predicted label from our model 6 hours ago, and $y_{t-6}$ the true label. Discrepancies between the predicted $\hat{y}_{t-6}$ and the label $y_{t-6}$ are fed into the the feed forward layers. Then the output of the feed forward layers will be fed into each RNN layer as a part of the input. We believe this can provide additional information about the correctness of previous hidden states. 

Note that, for the initial time steps $t \in \{1...6\}$, there is no feedback sent to the RNN from the previous time step, as we will need at least 6 hours of data to obtain the true label of AKI. In these cases, a default state is sent to each RNN layer instead. This default state is trained by backpropagation as neural network parameters.

\subsubsection{GRU and LSTM Fundamentals}

% \begin{figure}
% \centering
% \begin{subfigure}{.25\textwidth}
% \centering
% \includegraphics[width=.95\linewidth]{GRU_Cell}
% \caption{\label{fig:GRU_Cell}GRU Cell\cite{DBLP:journals/corr/ChungGCB14}}
% \end{subfigure}%
% \begin{subfigure}{.25\textwidth}
% \centering
% \includegraphics[width=.95\linewidth]{LSTM_Cell}
% \caption{\label{fig:LSTM_Cell}LSTM Cell\cite{DBLP:journals/corr/ChungGCB14}}
% \end{subfigure}
% \caption{Diagram of GRU and LSTM}
% \end{figure}
Gated Recurrent Unit (GRU) and Long Short-term Memory (LSTM) are the two most commonly used variants of RNN. They have been shown to work well on modeling sequential data with long-term dependencies.
% The structure of GRU is shown in Figure~\ref{fig:GRU_Cell}. The output ${h_t}$ is updated according to the following formula\cite{DBLP:journals/corr/ChungGCB14}:

% \begin{equation*}
% r_t = \sigma(W_rx_t + U_rh_{t-1})
% \end{equation*}
% \begin{equation*}
% \tilde{h}_t = \tanh(Wx_t+U(r_t\odot h_{t-1}))
% \end{equation*}
% \begin{equation*}
% z_t = \sigma(W_zx_t + U_zh_{t-1})
% \end{equation*}
% \begin{equation*}
% h_t = (1-z_t)\odot h_{t-1} + z_t\odot \tilde{h}_t
% \end{equation*}
% $r_t$ and $z_t$ denote the reset gate and update gate. And $\tilde{h}_t$ is the candidate of $h_t$.\\
% Similarly, LSTM Cell (demonstrated in Figure~\ref{fig:LSTM_Cell} has different gates to control the update from $h_{t-1}$ to $h_t$. It has three gates: input gate, output gate and forget gate as compared to GRU which has two gates. In addition to the output state $h_t$, LSTM Cell maintains a memory state $c_t$, which is for capturing the long-term dependencies. At each time step, the values of the gates are calculated and the output state and memory state are updated as follow\cite{hochreiter1997long,DBLP:journals/corr/ChungGCB14,DBLP:journals/corr/Chen16p}:

% \begin{equation*}
% i_t = \sigma(W_ix_t+U_ih_{t-1}+V_ic_{t-1})
% \end{equation*}
% \begin{equation*}
% f_t = \sigma(W_fx_t+U_fh_{t-1}+V_fc_{t-1})
% \end{equation*}
% \begin{equation*}
% o_t = \sigma(W_ox_t+U_oh_{t-1}+V_oc_{t-1})
% \end{equation*}
% \begin{equation*}
% \hat{c}_t = \tanh(W_cx_t+U_ch_{t-1})
% \end{equation*}
% \begin{equation*}
% c_t = f_tc_{t-1}+i_t\hat{c}_t
% \end{equation*}
% \begin{equation*}
% h_t = o_t\tanh(c_t)
% \end{equation*}

The common property shared between GRU and LSTM is the additive update process. The values of the gates depend on the input and the previous state. And update process is controlled by the gates together with the input and previous state. The function $h_t = RNN(h_{t-1}, x_t)$ performed by LSTM or GRU can therefore be divided into two steps: $gates = RNN_{gate}(h_{t-1}, x_t)$ and $h_t = RNN_{state}(gates, h_{t-1}, x_t)$. The joint distribution of a GRU or LSTM network factorizes as \cite{2016arXiv160507571F,DBLP:journals/corr/ChungKDGCB15}:
\begin{multline}
\\p(y_{1:T}, h_{1:T}, gate_{1:T} | x_{1:T}, h_0) \\ = \prod_{t=1}^T p(y_t|h_t)p(h_t|gate_t, h_{t-1}, x_t)p(gate_t|h_{t-1}, x_t)
\end{multline}
$h_0$ denotes the initial states of the GRU/LSTM Layers.\\

Note that these probability distributions modeled by the RNN are all deterministic, and that this is the joint distribution of a single-layer RNN. The joint distribution of multi-layer RNN factorizes into more components.

\subsubsection{Proposed Self-correcting Model}

As mentioned in section 3.3.1, at time step $t$, we will get the true label $y_{t-6}$ (i.e. whether the patient develops AKI). And we want this information to be used to improve the accuracy of the prediction at the current time step. Therefore, the joint probability modeled by the neural network should be:
\begin{multline}
\\p(\hat{y}_{1:T}, h_{1:T}, gate_{1:T} | x_{1:T}, h_0, y_{1:T-6}) \\
= \prod_{t=7}^T (p(\hat{y}_t|h_t)p(h_t|gate_t, h_{t-1}, x_t, \hat{y}_{t-6}, y_{t-6})\\
p(gate_t|h_{t-1}, x_t, \hat{y}_{t-6}, y_{t-6}))\\
\prod_{t=1}^6 p(\hat{y}_t|h_t)p(h_t|gate_t, h_{t-1}, x_t)\\
\end{multline}
$\hat{y}_{t}$ denotes the output of the neural network and $y_t$ denotes the label.\\

\begin{figure*}[h]
\centering
\includegraphics[width=0.95\linewidth]{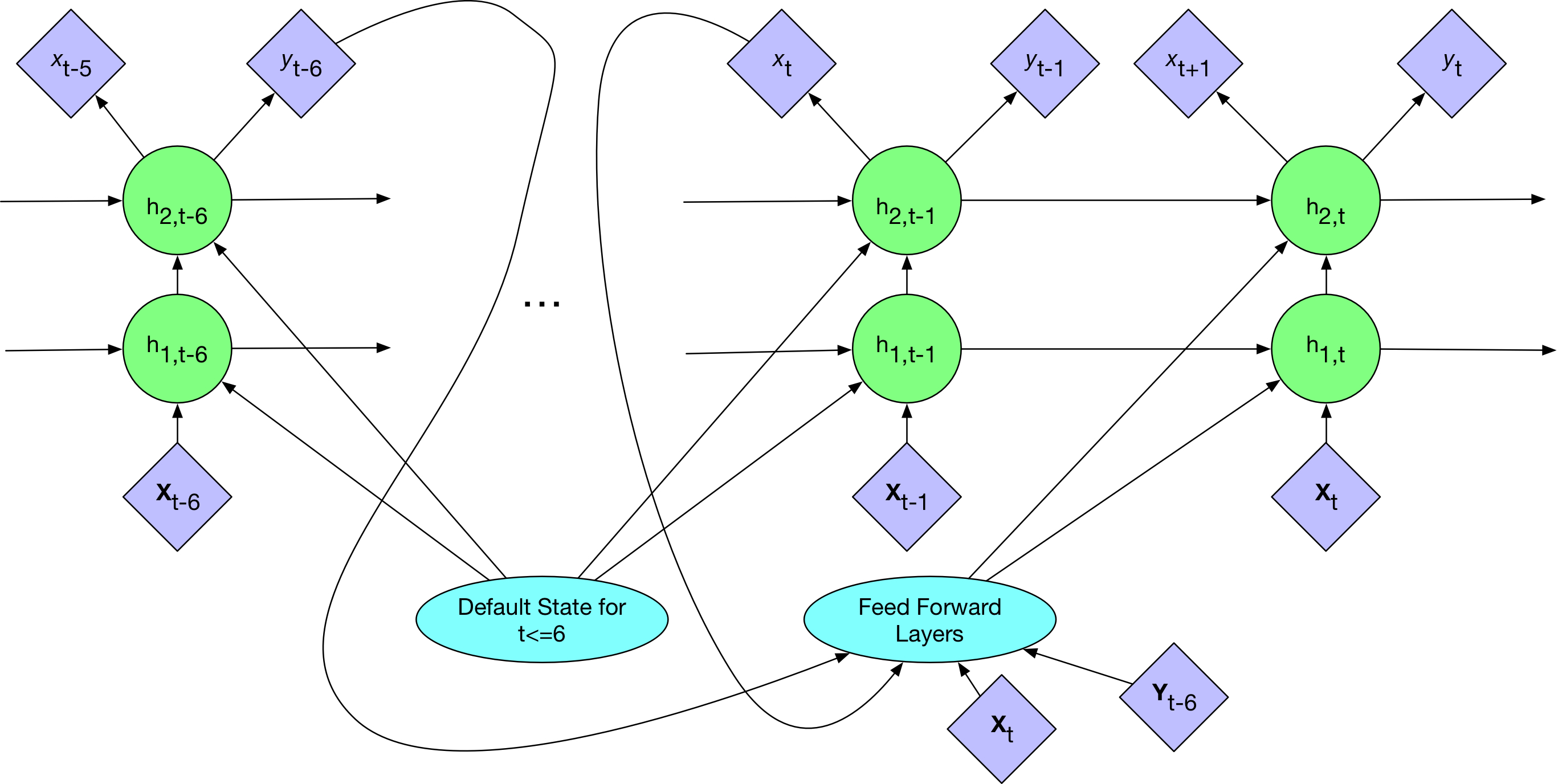}
\caption{RNN with self-correcting mechanism and regularization. Circles are used for RNN cells (either LSTM or GRU), while diamond shaped units are used for input and output. Italic letters (e.g. \textit{$x_t$, $y_t$}) denote the predicted values, while bold capital (e.g. \textbf{$X_t$, $Y_t$}) letters denote the actual values.}
\label{fig:self-correcting regularized rnn}
\end{figure*}
\subsubsection{Proposed Regularization for Self-correcting Model}

Note that for $t\in\{1...6\}$, the factorized joint distributions of Eq. (1) and (2) are the same. And the difference between Eq. (1) and (2) is that, in Eq. (2), the probability of $gate_t$ and $h_t$ is conditioned on $\hat{y}_{t-6}$ and $y_{t-6}$. Based on this probability model, we designed the neural network shown in Figure 2. Both $\hat{y}_{t-6}$ and $y_{t-6}$ are fed into each RNN layer. We call it "Self-correcting RNN" because the update of hidden state from $h_{t-1}$ to $h_t$ is based on the label $y_{t-6}$ and the predicted value $\hat{y}_{t-6}$ from the past time step. If the neural network makes a wrong prediction at the past time step, the hidden state is expected to be updated accordingly.
One challenge with the Self-correcting RNN is that the label and predicted value used to update $h_t$ are from 6 time steps ago. The model may achieve better performance if we can minimize the time gap. To further improve the Self-correcting RNN, we designed the regularization method for it. The Self-correcting RNN with regularization is shown in Figure~\ref{fig:self-correcting regularized rnn}. Instead of only predicting $\hat{y}_{t-1}$ at time step $t-1$, the model predicts $\hat{x}_{t}$ as well. Then at time step t, the predicted $\hat{x}_{t}$ and the input $x_{t}$ will be fed into the feed forward layers together with $\hat{y}_{t-6}$ and $y_{t-6}$. The joint probability distribution of this model is:
\begin{multline}
\\p(\hat{y}_{1:T}, \hat{x}_{2:T}, h_{1:T}, gate_{1:T} | x_{1:T}, h_0, y_{1:T-6}) \\
= p(\hat{y}_T|h_T)p(h_T|gate_T, h_{T-1}, \hat{x}_T, x_T, \hat{y}_{T-6}, y_{T-6})\\
p(gate_T|h_{T-1}, \hat{x}_T, x_T, \hat{y}_{T-6}, y_{T-6})\\
\prod_{t=7}^{T-1} (p(\hat{y}_t|h_t)p(\hat{x}_{t+1}|h_t)p(h_t|gate_t, h_{t-1}, \hat{x}_t, x_t, \hat{y}_{t-6}, y_{t-6})\\
p(gate_t|h_{t-1}, \hat{x}_t, x_t, \hat{y}_{t-6}, y_{t-6}))\\
\prod_{t=2}^6 (p(\hat{y}_t|h_t)p(\hat{x}_{t+1}|h_t)p(h_t|gate_t, h_{t-1}, \hat{x}_t, x_t))\\
p(\hat{y}_1|h_1)p(\hat{x}_2|h_1)p(h_1|gate_1, h_0, x_1)\\
\end{multline}
The factorized probability distribution Eq. (3) is more sophisticated than the one of the Self-correcting RNN without regularization Eq. (2). The main difference is that the probability distribution of $gate_t$ and $h_t$ now condition on $\hat{x}_t$ as well. When we train the model, we add the mean squared error between $\hat{x_t}$ and $x_t$ to the total loss after multiplying it with a certain coefficient. So the model will learn to predict $x_t$ by backpropagation. This regularization method boost the performance of the Self-correcting RNN in the following two ways:
\begin{enumerate}
\item It minimizes the time gap of the self-correcting mechanism.
\item It enforces the model to predict $x_{t+1}$ instead of only $y_t$. More information needs to be captured by the hidden state, and hence the variance of the model decreases.
\end{enumerate}

\subsubsection{Self-correcting RNN Model with Regularization for Regression}

All the models described above are classification models because the final result should be a binary value represents whether the patient will develop AKI in the next 6 hours. And the actual AKI label depends on whether the value of $urine \ output /  weight$ is larger than 0.5 mL/kg/h. So we also designed a Self-correcting RNN Model with Regularization for the urine output regression problem. The structure of this model is the similar to the one shown in Figure~\ref{fig:self-correcting regularized rnn}, except that it predicts the next-6-hour urine output instead of the AKI label, and then predicts the label based on the patient's weight and the predicted urine output. So it becomes a regression model. And mean square error is used for backpropagation. 

\subsubsection{Stop-gradient Technique for Feedback Loop}
\begin{figure}[h]
\centering
\includegraphics[width=0.9\linewidth]{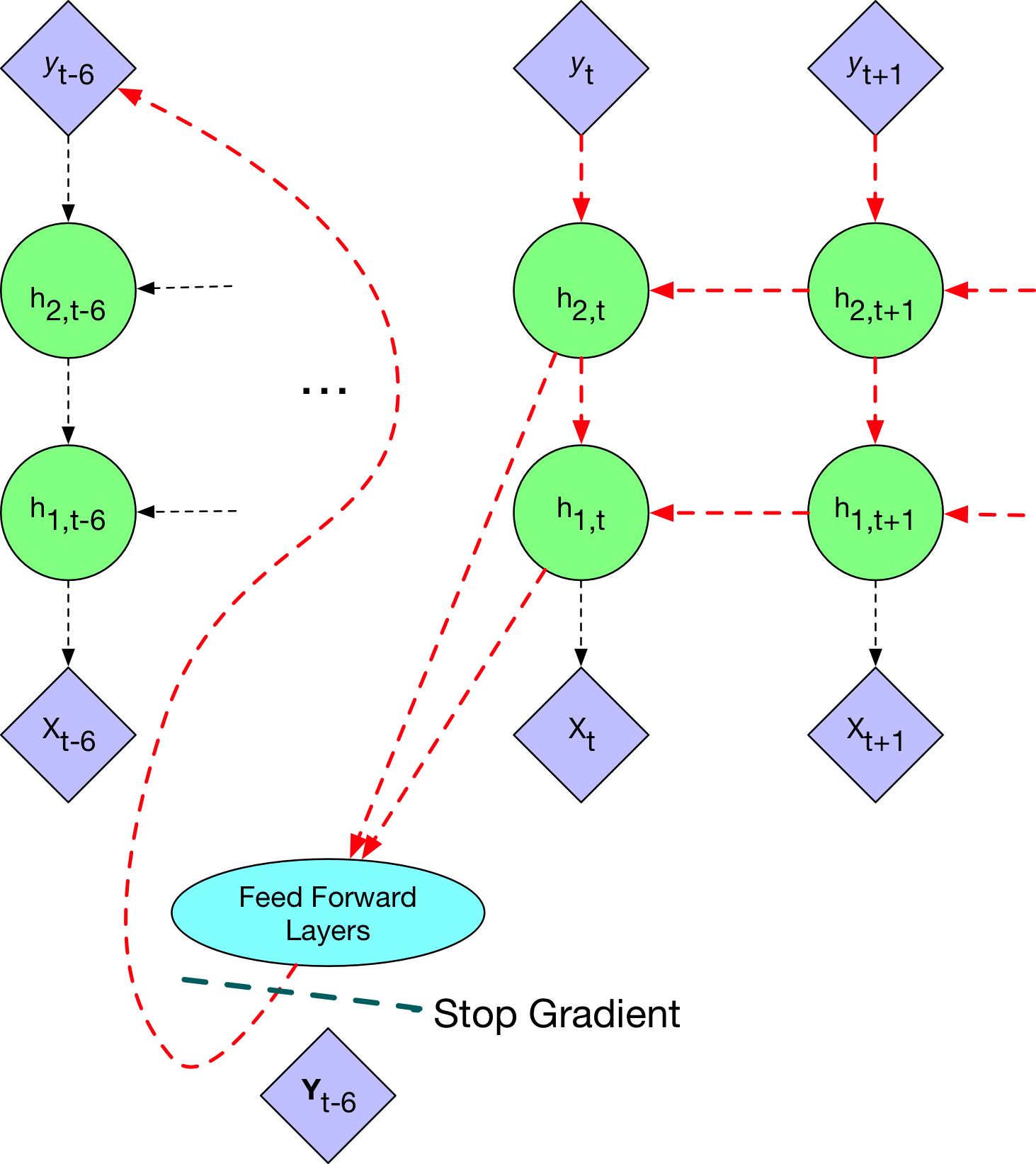}
\caption{\label{fig:Self-correcting_RNN_Gradient}Diagram of backpropagation of the error. Dotted lines indicate the direction of backpropagation, and the dotted lines in red show how the errors in the future time steps are backpropagated to $\hat{y}$ through the feedback loop connection. The green dotted line indicates where we apply the stop-gradient technique.}
\end{figure}

Another challenge  with the proposed self-correcting models is that the gradient of $y_t$ is affected by the errors at the future time steps. Let $J(\theta)$ denote the cost function of the parameters $\theta$ of the neural network. When we train the RNN models using gradient descent algorithm, the partial derivative $\frac{\partial J}{\partial \hat{y}_t}$ is first calculated and then backpropagated through the time. In the traditional RNN, the partial derivative of $\hat{y}_t$ is $\frac{\partial J}{\partial \hat{y}_t} = \frac{\partial J_t}{\partial \hat{y}_t}$\cite{DBLP:journals/corr/Chen16p}, where $J_t$ is the cross-entropy loss between $\hat{y_t}$ and the label $y_t$. The partial derivative of $y_t$ is not affected by the loss at the other time steps. This is the desired property of the RNN. In our Self-correcting RNN model, the partial derivative of $y_t$ is:
\begin{equation*}
\frac{\partial J}{\partial \hat{y}_t} = \frac{\partial J_t}{\partial \hat{y}_t} + \sum_{i=6}^{T-t-6}\frac{\partial J_{t+i}}{\partial \hat{y}_t}
\end{equation*}
This is because the loss at the future time step can be backpropagated through the RNN layers and feed forward layers, and finally to ${y_t}$, as shown in Figure~\ref{fig:Self-correcting_RNN_Gradient}. And this is not what we desired. Intuitively, the problem is that output layer does not only need to predict $\hat{y}_t$ accurately, but also need to generate the $\hat{y}_t$ such that the value will later lead to more accurate $\hat{y}_{t+6}$ given the current neural network parameters, since $\frac{\partial J_{t+6}}{\partial \hat{y}_t}$ is also a component of $\frac{\partial J}{\partial \hat{y}_t}$. This is an undesired property and may potentially affect the performance of the model. The $\hat{y}_t$ predicted by the model should only be used for the self-correcting mechanism to boost the prediction accuracy at future time step. \\
To tackle this issue, we truncate the gradient right before feeding the feedback into the feedback network, as shown in Figure~\ref{fig:Self-correcting_RNN_Gradient}. This is referred to as the stop-gradient technique for the self-correcting models. 

%%%%%%%%%%%%%%%%%%%%%%%%%%%%%%%%%%%%%%%%%%%%%%

%\begin{table*}
%\begin{center}
%  \begin{tabular}{||c c c c c||} 
%  \hline
%  Model & MIMIC & MIMIC$_{t>6}$ & eICU & eICU$_{t>6}$ \\ [0.5ex] 
%  \hline\hline
%  Multi-layer GRU & 0.741 &0.743 & 0.770  & 0.812\\
%  \hline
%  Self-correcting RNN & 0.879 &0.889 & 0.789 & 0.837\\
%  \hline
%  Self-correcting Regression with Regularization & 0.878 &0.886 & 0.827 &0.861\\ [1ex] 
%  \hline
%  Self-correcting RNN with Regularization & \textbf{0.896} & \textbf{0.893} & \textbf{0.851}& \textbf{0.871}\\
%  \hline
% \end{tabular}
% \end{center}
% \caption{AUC of the four models on MIMIC-III and eICU over all time steps.}
% \label{tab:result1}
% \end{table*}

\section{Results and Discussions}
\subsection{Experiment Setup}
Systematic experiments were conducted to compare the performance of our proposed models and the previously proposed RNN approaches. 
%And we want to evaluate whether the self-correcting mechanism helps to boost the performance of RNN. 
The MIMIC-III and eICU datasets were chosen to validate our proposed methods as they are representative datasets with the richest critical care EHR data. Our models were trained and tested on these two datasets separately.
%
% We selected 2 static features and 21 dynamic features to be the input of the data models. These features include the important vital signs, laboratory test results, medication records and medical intervention. Some of the data, like heart rate, are collected more than once per hour for every patients. And some data is collected daily. We decided to transform the data into time series format where the time gap between two data point is 1 hour. The time steps when there is no recorded data are filled with the weighted average value of the nearest data points.
%

In this study, we only included patients who stayed in ICU for at least 12 hours. This criterion was set based on two considerations: (1) patients, who were discharged or died within the first 12 hours of ICU stay, were very unique patients that do not fit our clinical application, and (2) our proposed self-correcting mechanism only starts from $t=7$ onwards. In addition, we have also removed patients whose data for the selected variables was not recorded for at least once during the ICU stay. 
With these inclusion and exclusion criteria, we ended up with about 25,000 patients out of the 40,000 MIMIC patients and about 11,000 patients out of 20,0859 eICU patients. 
We also eliminated the outliers in the extracted time-series data (e.g. negative heart rate, unreasonably high body temperature) by removing the extreme data points at the upper or lower one percentiles. 

For each of the two datasets, we trained four models: the proposed Self-correcting RNN, the proposed Self-correcting RNN with Regularization, the proposed Self-correcting regression RNN with Regularization and normal multi-layer GRU network. The normal multi-layer GRU network is chosen to be the baseline model. For all these four RNN models, we built the models with the same architecture: two layers of GRU, 128 neurons in each layer. We applied the same settings for dropout, gradient clipping and re-weighted loss function to address the imbalanced dataset. All the four models converged. In order to encourage reproducibility of research, the source code for this study is released online to public. 

\subsection{Performance Results}

\begin{table*}[htb]
\begin{center}
 \begin{tabular}{||c c c c c||} 
 \hline
 Model & MIMIC (testing) & MIMIC (training) & eICU (testing) & eICU (training) \\ [0.5ex] 
 \hline\hline
 Multi-layer GRU (baseline)  &0.743  &0.777  & 0.812 & 0.836\\
 \hline
 \textbf{Self-correcting RNN*}   &0.889 &0.892  & 0.837 & 0.875\\
 \hline
 \textbf{Self-correcting Regression with Regularization*}   &0.886 &0.891  &0.861 &0.894\\ [1ex] 
 \hline
 \textbf{Self-correcting RNN with Regularization*}  & 0.893 & 0.897 & 0.871 & 0.899\\
 \hline
\end{tabular}
\end{center}
\caption{AUC of the four models on MIMIC-III and eICU. * Proposed methods.}
\label{tab:result1}
\end{table*}

\begin{figure*}[htb]
\centering
\begin{tabular}{cc}
\includegraphics[width=0.5\linewidth]{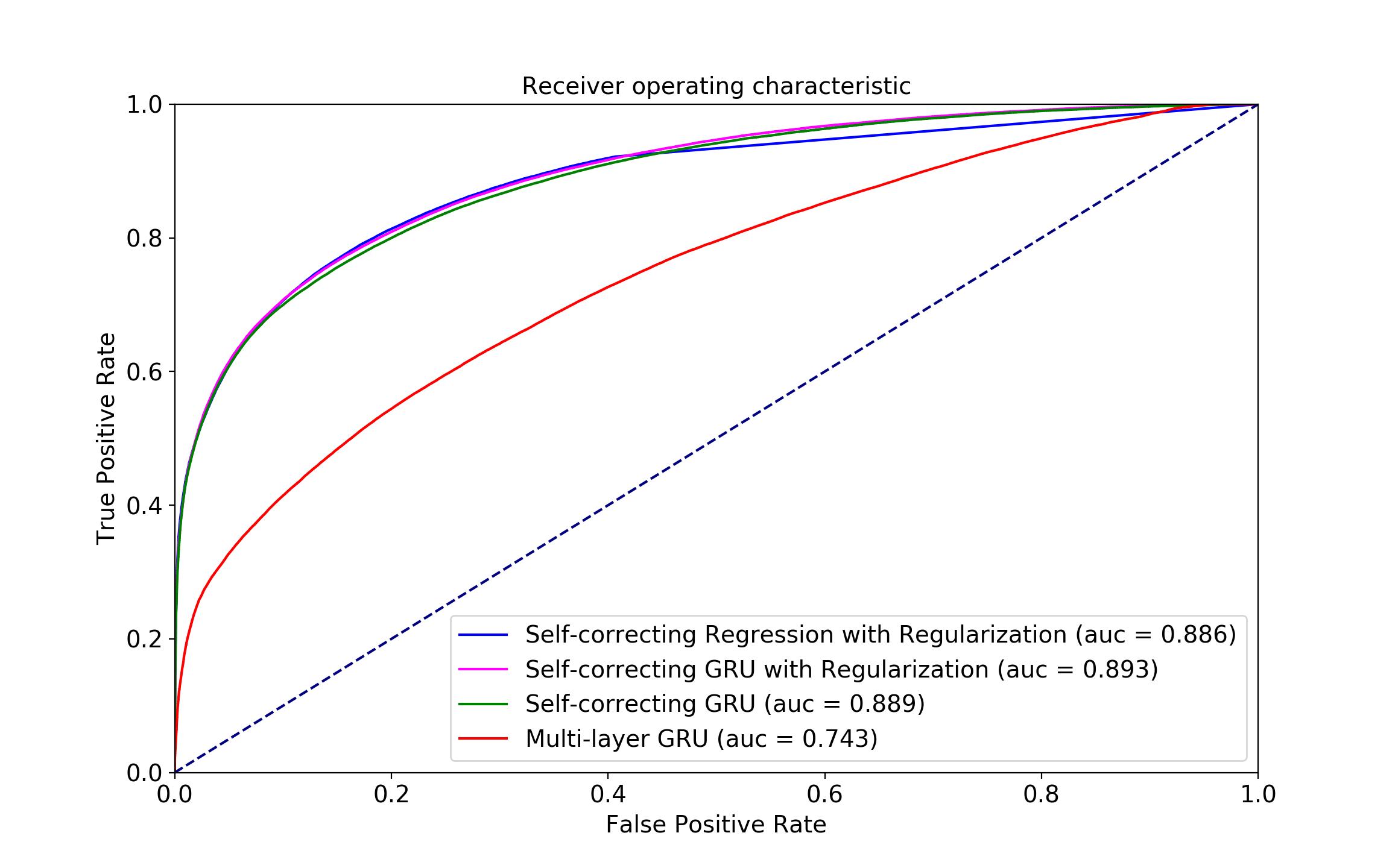}
&
\includegraphics[width=0.5\linewidth]{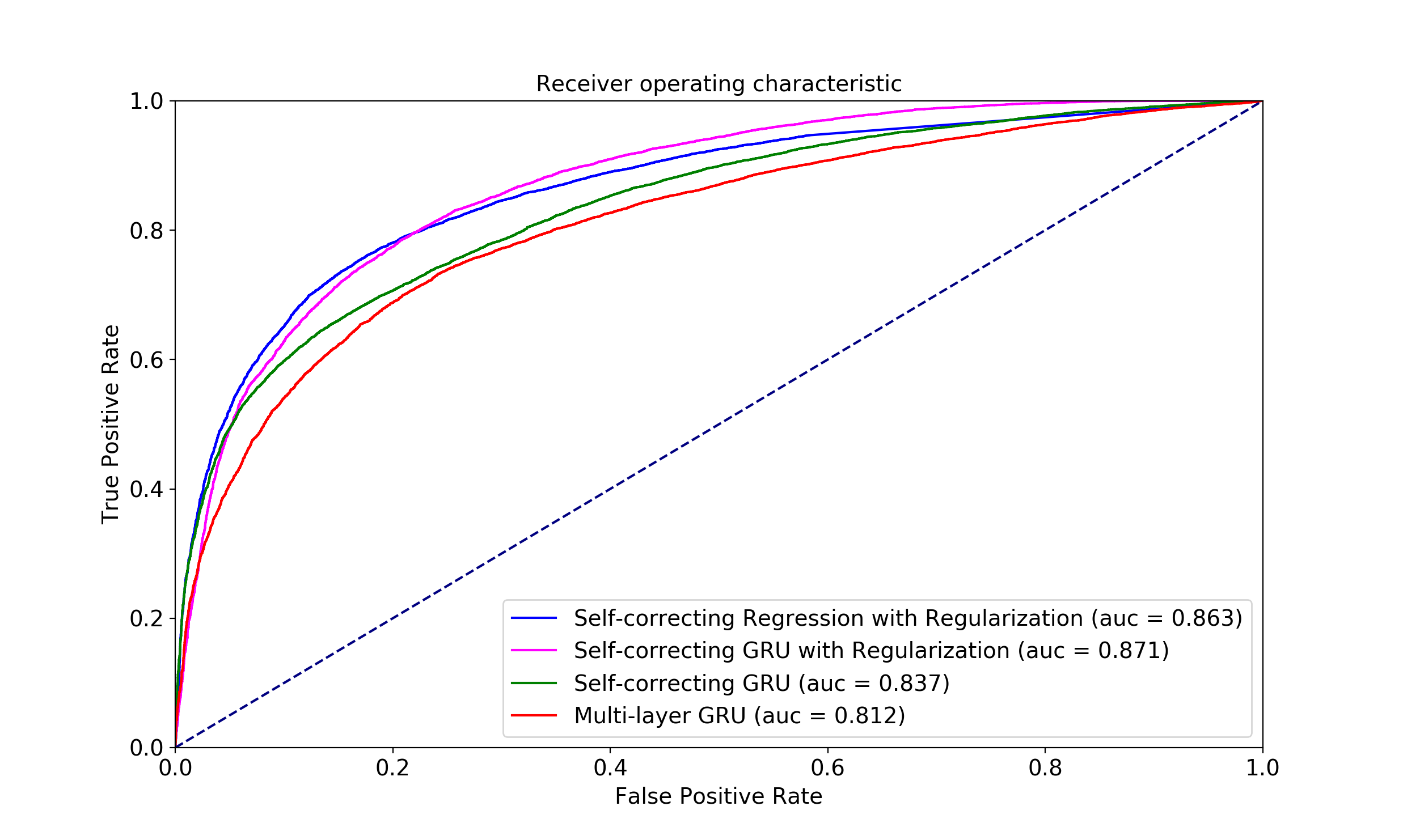} \\
(a) & (b)
\end{tabular}
\caption{ROC of the four comparing models (a) on the MIMIC-III dataset for time steps $t>6$, and (b) on the eICU dataset for time steps $t>6$. The red line indicates the baseline RNN model, and the other lines are ones for our proposed methods.}
\label{fig:mimic_roc}
\end{figure*}

For each dataset, we measured the performance of the models based on area under ROC curve (AUC). We only calculated the AUC for time steps $t>6$, as our proposed self-correcting mechanism starts at $t=7$. 
The ROC curves of all the four models are shown in Figure~\ref{fig:mimic_roc}, and the corresponding AUC were reported in Table~\ref{tab:result1}.

% The ROC curves of all the four models trained with MIMIC-III dataset are shown in Figure~\ref{fig:mimic_roc} and Figure~\ref{fig:mimic_roc}. The ROC curves of all the four models trained with eICU dataset are shown in Figure~\ref{fig:mimic_roc} and Figure~\ref{fig:mimic_roc}. And the corresponding AUC of the models are shown in Table~\ref{tab:result1}, Table~\ref{tab:result2}, Table~\ref{tab:result3} and Table~\ref{tab:result4}. 

All our proposed models outperform the baseline RNN model over both the MIMIC and eICU dataset. The Self-correcting RNN with regularization achieves the highest accuracy and AUC among the proposed models.

\subsection{The Benefits of Self-correcting Mechanism}

All the three self-correcting models outperformed the traditional multi-layer GRU. All the models achieved better results on MIMIC-III dataset because of the larger size of the dataset and better quality of the data. From the results on MIMIC-III dataset, there is a huge gap between the AUC of Self-correcting RNN and the traditional multi-layer GRU. This is because the additional information provided by the feedback network is helpful for the RNN update process. These results verified our hypothesis that the self-correcting mechanism can boost our model's performance.

% \begin{figure*}
% \centering
% \begin{tabular}{cc}
% \includegraphics[width=0.5\linewidth]{mimic_roc}
% &
% \includegraphics[width=0.5\linewidth]{mimic_after6_roc} \\
% (a) & (b) \\
% \includegraphics[width=0.5\linewidth]{eicu_roc}
% &
% \includegraphics[width=0.5\linewidth]{eicu_after6h_roc} \\
% (c) & (d) 
% \end{tabular}
% \caption{ROC of the four comparing models on (a) MIMIC-III dataset, (b) on MIMIC-III dataset for time steps $t>6$.}
% \label{fig:mimic_roc}
% \end{figure*}

\subsection{The Benefits of the Proposed Regularization Method}

Self-correcting RNN with Regularization achieved the highest AUC. And on eICU dataset, the Self-correcting Regression RNN with Regularization also achieved much higher AUC than the Self-correcting RNN model. It indicates that the proposed regularization method helps to further improve the performance of the models by enforcing the model to predict the future input. And as shown in Table~\ref{tab:result1}, our experiment on eICU dataset verifies that the Self-correcting RNN model with regularization has a smaller performance gap between training and testing data, as compared to the one without regularization. It proves that the regularization method reduces the variance of the model.

\subsection{The Benefit of the Stopped Gradient for the Feedback Neural}

To verify that the benefits from the proposed stop-gradient technique, we trained a Self-correcting RNN with Regularization without applying the stop-gradient technique. The AUC of this model was 0.852, while the AUC was 0.896 with the stop-gradient technique. The difference in the AUC verified that the stop-gradient technique is critical to the self-correcting models, because it prevent the gradient of $y_t$ from being affected by the future errors.

\section{Conclusions and Future Work}
We proposed a novel self-correcting enhancement to RNNs to better predict the onset of acute conditions in ICU. 
%With the proposed regularization step, our proposed self-correcting mechanism takes into account of not only its previous prediction error on the onset labels but also its estimation errors on the input data.
The proposed self-correcting mechanism made the update process of the hidden state of RNN to be dependent on the previous predicted output and the corresponding label. The additional information provided by the previous predicted output and label helped to boost the performance of model. We also proposed a regularization method for our model, which takes into account not only the model's prediction error on the label but also its estimation errors on the input data. The regularization method reduces the variance of the model, and also reduces the time gap for self-correcting mechanism. The method we proposed can be apply on both classification and regression models.
Our proposed models were tested on real-world large scale ICU dataset MIMIC-III and eICU and were shown to constantly outperform the baseline multi-layer GRU model.
Moreover, although we focus on the prediction of acute kidney injury as an example, the proposed model can be easily generalized to predict the other acute conditions in ICU. 

This is the first phase of our project. Inspired by the achieved promising results, we plan to move on further validate the proposed algorithm locally at our own hospital with the ultimate goal to deploy it as an decision support tool.

\clearpage
\newpage
\bibliography{bioaai}

\begin{thebibliography}{}

\bibitem[\protect\citeauthoryear{Association}{2004}]{american2004international}
Association, A.~M.
\newblock 2004.
\newblock {\em International classification of diseases, 9th revision, clinical
  modification: physician ICD-9-CM, 2005: volumes 1 and 2, color-coded,
  illustrated}, volume~1.
\newblock Amer Medical Assn.

\bibitem[\protect\citeauthoryear{Bellomo \bgroup et al\mbox.\egroup
  }{2004}]{bellomo2004acute}
Bellomo, R.; Ronco, C.; Kellum, J.~A.; Mehta, R.~L.; and Palevsky, P.
\newblock 2004.
\newblock Acute renal failure--definition, outcome measures, animal models,
  fluid therapy and information technology needs: the second international
  consensus conference of the acute dialysis quality initiative (adqi) group.
\newblock {\em Critical care} 8(4):R204.

\bibitem[\protect\citeauthoryear{Brown \bgroup et al\mbox.\egroup
  }{2007}]{brown2007multivariable}
Brown, J.~R.; Cochran, R.~P.; Leavitt, B.~J.; Dacey, L.~J.; Ross, C.~S.;
  MacKenzie, T.~A.; Kunzelman, K.~S.; Kramer, R.~S.; Hernandez~Jr, F.; Helm,
  R.~E.; et~al.
\newblock 2007.
\newblock Multivariable prediction of renal insufficiency developing after
  cardiac surgery.
\newblock {\em Circulation} 116(11\_supplement):I--139.

\bibitem[\protect\citeauthoryear{Chen}{2016}]{DBLP:journals/corr/Chen16p}
Chen, G.
\newblock 2016.
\newblock A gentle tutorial of recurrent neural network with error
  backpropagation.
\newblock {\em CoRR} abs/1610.02583.

\bibitem[\protect\citeauthoryear{Cheng \bgroup et al\mbox.\egroup
  }{2016}]{cheng2016risk}
Cheng, Y.; Wang, F.; Zhang, P.; and Hu, J.
\newblock 2016.
\newblock Risk prediction with electronic health records: A deep learning
  approach.
\newblock In {\em Proceedings of the 2016 SIAM International Conference on Data
  Mining},  432--440.
\newblock SIAM.

\bibitem[\protect\citeauthoryear{Chertow \bgroup et al\mbox.\egroup
  }{2005}]{chertow2005acute}
Chertow, G.~M.; Burdick, E.; Honour, M.; Bonventre, J.~V.; and Bates, D.~W.
\newblock 2005.
\newblock Acute kidney injury, mortality, length of stay, and costs in
  hospitalized patients.
\newblock {\em Journal of the American Society of Nephrology}
  16(11):3365--3370.

\bibitem[\protect\citeauthoryear{Cho \bgroup et al\mbox.\egroup
  }{2014}]{cho2014learning}
Cho, K.; Van~Merri{\"e}nboer, B.; Gulcehre, C.; Bahdanau, D.; Bougares, F.;
  Schwenk, H.; and Bengio, Y.
\newblock 2014.
\newblock Learning phrase representations using rnn encoder-decoder for
  statistical machine translation.
\newblock {\em arXiv preprint arXiv:1406.1078}.

\bibitem[\protect\citeauthoryear{Choi \bgroup et al\mbox.\egroup
  }{2016}]{choi2016doctor}
Choi, E.; Bahadori, M.~T.; Schuetz, A.; Stewart, W.~F.; and Sun, J.
\newblock 2016.
\newblock Doctor ai: Predicting clinical events via recurrent neural networks.
\newblock In {\em Machine Learning for Healthcare Conference},  301--318.

\bibitem[\protect\citeauthoryear{Chung \bgroup et al\mbox.\egroup
  }{2015}]{DBLP:journals/corr/ChungKDGCB15}
Chung, J.; Kastner, K.; Dinh, L.; Goel, K.; Courville, A.~C.; and Bengio, Y.
\newblock 2015.
\newblock A recurrent latent variable model for sequential data.
\newblock {\em CoRR} abs/1506.02216.

\bibitem[\protect\citeauthoryear{Cruz12 \bgroup et al\mbox.\egroup
  }{}]{cruz12early}
Cruz12, H.; Grasnick, B.; Dinger, H.; Bier, F.; and Meinel, C.
\newblock Early detection of acute kidney injury with bayesian networks.

\bibitem[\protect\citeauthoryear{{Fraccaro} \bgroup et al\mbox.\egroup
  }{2016}]{2016arXiv160507571F}
{Fraccaro}, M.; {Kaae S{\o}nderby}, S.; {Paquet}, U.; and {Winther}, O.
\newblock 2016.
\newblock {Sequential Neural Models with Stochastic Layers}.
\newblock {\em ArXiv e-prints}.

\bibitem[\protect\citeauthoryear{Harty}{2014}]{Harty2014Prevention}
Harty, J.
\newblock 2014.
\newblock Prevention and management of acute kidney injury.
\newblock {\em Ulster Medical Journal} 83(3):149.

\bibitem[\protect\citeauthoryear{Hochreiter and
  Schmidhuber}{1997}]{hochreiter1997long}
Hochreiter, S., and Schmidhuber, J.
\newblock 1997.
\newblock Long short-term memory.
\newblock {\em Neural computation} 9(8):1735--1780.

\bibitem[\protect\citeauthoryear{Johnson \bgroup et al\mbox.\egroup
  }{2016}]{johnson2016mimic}
Johnson, A.~E.; Pollard, T.~J.; Shen, L.; Li-wei, H.~L.; Feng, M.; Ghassemi,
  M.; Moody, B.; Szolovits, P.; Celi, L.~A.; and Mark, R.~G.
\newblock 2016.
\newblock Mimic-iii, a freely accessible critical care database.
\newblock {\em Scientific data} 3:160035.

\bibitem[\protect\citeauthoryear{Kate \bgroup et al\mbox.\egroup
  }{2016}]{kate2016prediction}
Kate, R.~J.; Perez, R.~M.; Mazumdar, D.; Pasupathy, K.~S.; and Nilakantan, V.
\newblock 2016.
\newblock Prediction and detection models for acute kidney injury in
  hospitalized older adults.
\newblock {\em BMC medical informatics and decision making} 16(1):39.

\bibitem[\protect\citeauthoryear{Lipton \bgroup et al\mbox.\egroup
  }{2015}]{lipton2015learning}
Lipton, Z.~C.; Kale, D.~C.; Elkan, C.; and Wetzel, R.
\newblock 2015.
\newblock Learning to diagnose with lstm recurrent neural networks.
\newblock {\em arXiv preprint arXiv:1511.03677}.

\bibitem[\protect\citeauthoryear{Ma, Xiao, and Wang}{2018}]{ma2018health}
Ma, T.; Xiao, C.; and Wang, F.
\newblock 2018.
\newblock Health-atm: A deep architecture for multifaceted patient health
  record representation and risk prediction.
\newblock In {\em Proceedings of the 2018 SIAM International Conference on Data
  Mining},  261--269.
\newblock SIAM.

\bibitem[\protect\citeauthoryear{Miotto \bgroup et al\mbox.\egroup
  }{2017}]{miotto2017deep}
Miotto, R.; Wang, F.; Wang, S.; Jiang, X.; and Dudley, J.~T.
\newblock 2017.
\newblock Deep learning for healthcare: review, opportunities and challenges.
\newblock {\em Briefings in bioinformatics}.

\bibitem[\protect\citeauthoryear{Mohamadlou \bgroup et al\mbox.\egroup
  }{2018}]{mohamadlou2018prediction}
Mohamadlou, H.; Lynn-Palevsky, A.; Barton, C.; Chettipally, U.; Shieh, L.;
  Calvert, J.; Saber, N.~R.; and Das, R.
\newblock 2018.
\newblock Prediction of acute kidney injury with a machine learning algorithm
  using electronic health record data.
\newblock {\em Canadian Journal of Kidney Health and Disease}
  5:2054358118776326.

\bibitem[\protect\citeauthoryear{Moody, Mark, and
  Goldberger}{2001}]{moody2001physionet}
Moody, G.~B.; Mark, R.~G.; and Goldberger, A.~L.
\newblock 2001.
\newblock Physionet: a web-based resource for the study of physiologic signals.
\newblock {\em IEEE Engineering in Medicine and Biology Magazine} 20(3):70--75.

\bibitem[\protect\citeauthoryear{Palomba \bgroup et al\mbox.\egroup
  }{2007}]{palomba2007acute}
Palomba, H.; De~Castro, I.; Neto, A.; Lage, S.; and Yu, L.
\newblock 2007.
\newblock Acute kidney injury prediction following elective cardiac surgery:
  Akics score.
\newblock {\em Kidney international} 72(5):624--631.

\bibitem[\protect\citeauthoryear{Pham \bgroup et al\mbox.\egroup
  }{2016}]{pham2016deepcare}
Pham, T.; Tran, T.; Phung, D.; and Venkatesh, S.
\newblock 2016.
\newblock Deepcare: A deep dynamic memory model for predictive medicine.
\newblock In {\em Pacific-Asia Conference on Knowledge Discovery and Data
  Mining},  30--41.
\newblock Springer.

\bibitem[\protect\citeauthoryear{Qiao \bgroup et al\mbox.\egroup
  }{2018}]{qiao2018pairwise}
Qiao, Z.; Zhao, S.; Xiao, C.; Li, X.; Qin, Y.; and Wang, F.
\newblock 2018.
\newblock Pairwise-ranking based collaborative recurrent neural networks for
  clinical event prediction.
\newblock In {\em IJCAI},  3520--3526.

\bibitem[\protect\citeauthoryear{Sutherland \bgroup et al\mbox.\egroup
  }{2016}]{sutherland2016utilizing}
Sutherland, S.~M.; Chawla, L.~S.; Kane-Gill, S.~L.; Hsu, R.~K.; Kramer, A.~A.;
  Goldstein, S.~L.; Kellum, J.~A.; Ronco, C.; and Bagshaw, S.~M.
\newblock 2016.
\newblock Utilizing electronic health records to predict acute kidney injury
  risk and outcomes: workgroup statements from the 15 th adqi consensus
  conference.
\newblock {\em Canadian journal of kidney health and disease} 3(1):11.

\bibitem[\protect\citeauthoryear{Thakar \bgroup et al\mbox.\egroup
  }{2005}]{thakar2005clinical}
Thakar, C.~V.; Arrigain, S.; Worley, S.; Yared, J.-P.; and Paganini, E.~P.
\newblock 2005.
\newblock A clinical score to predict acute renal failure after cardiac
  surgery.
\newblock {\em Journal of the American Society of Nephrology} 16(1):162--168.

\bibitem[\protect\citeauthoryear{Weisenthal \bgroup et al\mbox.\egroup
  }{2017}]{weisenthal2017sum}
Weisenthal, S.; Liao, H.; Ng, P.; and Zand, M.
\newblock 2017.
\newblock Sum of previous inpatient serum creatinine measurements predicts
  acute kidney injury in rehospitalized patients.
\newblock {\em arXiv preprint arXiv:1712.01880}.

\end{thebibliography}
\bibliographystyle{aaai}
\end{document}